\documentclass[10pt,twocolumn,letterpaper]{article}
\pdfoutput=1
\usepackage{iccv}
\usepackage{times}
\usepackage{epsfig}
\usepackage{graphicx}
\usepackage{amsmath}
\usepackage{amssymb}
\usepackage{multirow}
\usepackage[table,xcdraw]{xcolor}

\usepackage{pgfplots}
\pgfplotsset{compat=1.18}
\usepackage{tikz}
\usepackage{times}
\usetikzlibrary{positioning}
\usetikzlibrary{calc}
\usetikzlibrary{shapes.geometric}

\usepackage{enumitem}

\usepackage[pagebackref=true,breaklinks=true,letterpaper=true,colorlinks,bookmarks=false]{hyperref}
\iccvfinalcopy

\def\iccvPaperID{4310}

\ificcvfinal\fi

\begin{document}

\title{Accurate and Fast Compressed Video Captioning}

\author{
    Yaojie Shen$^{1,2*}$, 
    Xin Gu$^{1,2*}$, 
    Kai Xu$^3$, 
    Heng Fan$^4$,
    Longyin Wen$^3$, 
    Libo Zhang$^{1,2\dagger}$\\
    $^1$ Institute of Software, Chinese Academy of Sciences, Beijing, China\\
    $^2$ University of Chinese Academy of Sciences, Beijing, China\\
    $^3$ ByteDance Inc., San Jose, USA\\
    $^4$ Department of Computer Science and Engineering, University of North Texas, Denton TX, USA
}

\maketitle

\def\thefootnote{*}\footnotetext{The two authors make equal contributions and are co-first authors. }
\def\thefootnote{\textdagger}\footnotetext{Corresponding author: Libo Zhang (libo@iscas.ac.cn).}
\def\thefootnote{\arabic{footnote}}

\ificcvfinal\thispagestyle{empty}\fi

\begin{abstract}

Existing video captioning approaches typically require to first sample video frames from a decoded video and then conduct a subsequent process (\eg, feature extraction and/or captioning model learning). In this pipeline, manual frame sampling may ignore key information in videos and thus degrade performance. Additionally, redundant information in the sampled frames may result in low efficiency in the inference of video captioning. Addressing this, we study video captioning from a different perspective in compressed domain, which brings multi-fold advantages over the existing pipeline: 1) Compared to raw images from the decoded video, the compressed video, consisting of I-frames, motion vectors and residuals, is highly distinguishable, which allows us to leverage the entire video for learning without manual sampling through a specialized model design; 2) The captioning model is more efficient in inference as smaller and less redundant information is processed. We propose a simple yet effective end-to-end transformer in the compressed domain for video captioning that enables learning from the compressed video for captioning. We show that even with a simple design, our method can achieve state-of-the-art performance on different benchmarks while running almost 2$\times$ faster than existing approaches. Code is available at \url{https://github.com/acherstyx/CoCap}.

\end{abstract}

\section{Introduction}

Video captioning is a representative example of applying deep learning to the fields of computer vision and natural language processing with a long list of applications, such as blind navigation, video event commentary, and human-computer interaction. To generate captions for a video, the model needs to not only identify objects and actions in the video, but also be able to express them accurately in natural language. Despite significant progress, accurate and fast video captioning remains a challenge.

\begin{figure}[tbp]
    \centering
    \includegraphics[width=0.99\linewidth]{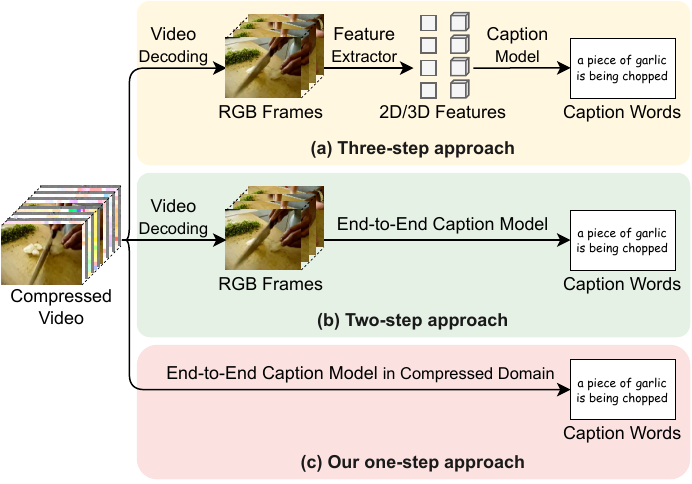}
    \caption{Comparing our method with prior methods for video captioning. Prior works are all based on decoding video frames. The difference between them is that some methods use offline extracted multiple features as input and generate captions, while others directly take dense video frames as input. By avoiding heavy redundant information and offline multiple feature extraction, our method speedup the caption generation process while maintaining high quality results.}
    \label{fig:compare_with_prior_method}
\end{figure}

\begin{figure}[tbp]
\centering
\begin{tikzpicture}[inner sep=0pt]
\scriptsize
\pgfplotsset{
  log x ticks with fixed point/.style={
      xticklabel={
        \pgfkeys{/pgf/fpu=true}
        \pgfmathparse{exp(\tick)}
        \pgfmathprintnumber[fixed relative, precision=1]{\pgfmathresult}
        \pgfkeys{/pgf/fpu=false}
      }
  },
  log y ticks with fixed point/.style={
      yticklabel={
        \pgfkeys{/pgf/fpu=true}
        \pgfmathparse{exp(\tick)}
        \pgfmathprintnumber[fixed relative, precision=1]{\pgfmathresult}
        \pgfkeys{/pgf/fpu=false}
      }
  },
  compat=newest
}

\pgfdeclareplotmark{mystar}{
    \node[star,star point ratio=2.25,minimum size=6pt,inner sep=0pt,draw=black,solid,fill=red] {};
}
\pgfdeclareplotmark{mysquare}{
    \node[circle,minimum size=5pt,inner sep=0pt,draw=black,solid,fill=blue] {};
}

\begin{axis}[
    xmin=0.1, xmax=11,
    ymin=48, ymax=59,
    xmode=log,
    log x ticks with fixed point,
    axis x line=bottom,
    axis y line=left,
    xlabel=Second,
    ylabel=CIDEr,
    x label style={at={(axis description cs:0.5,-0.13)},anchor=north,font=\small},
    y label style={at={(axis description cs:-0.1,0.5)},anchor=south,font=\small},
    ytick={48,50,52,54,56,58},
    xticklabel style={below=0.2em},
    yticklabel style={left=0.2em},
    grid=both,
    grid style={line width=.5pt,draw=black!70,dotted,step=0.1},
    every major tick/.append style={thick, major tick length=3pt, black},
    every minor tick/.append style={line width=.5pt, minor tick length=0pt, gray!40},
    tick align=outside,
    tickpos=left,
    width=\linewidth,
    height=\linewidth/3*2,
    axis line style = {thick},
];
\addplot[
    only marks,
    mark=mysquare,mark options={color=orange},
    mark size=2pt,
    visualization depends on=\thisrow{alignment} \as \alignment,
    visualization depends on=\thisrow{xshift} \as \xshift,
    nodes near coords, 
    point meta=explicit symbolic,
    every node near coord/.style={anchor=\alignment, xshift=\xshift, yshift=-0.4, color=black}
] table [
    meta index=2
]{
    Time        acc     name                                        alignment   xshift
    0.344       53.8    SwinBERT~\cite{SwinBERT}                    0           49
    2.709       51.5    HMN~\cite{HMN}                              0           33
    0.577       49.5    SGN~\cite{SGN}                              0           31
};
\addplot[
    only marks,
    mark=mystar,mark options={color=pink},
    mark size=2pt,
    visualization depends on=\thisrow{alignment} \as \alignment,
    visualization depends on=\thisrow{xshift} \as \xshift,
    nodes near coords,
    point meta=explicit symbolic,
    every node near coord/.style={anchor=\alignment, xshift=\xshift, yshift=-0.4, font=\bfseries, color=black}
] table [
    meta index=2
]{
    Time        acc     name                        alignment   xshift
    0.238       57.70   Ours(I+MV+Res,~L-14)        0           75
    0.178       56.20   Ours(I+MV+Res)              0           58
    0.153       55.30   Ours(I+MV)                  0           43
    0.146       54.10   Ours(I)                     0           27
};
\draw[red] (0.238,57.70)  -- (0.178,56.20);
\draw[red] (0.178,56.20)  -- (0.153,55.30);
\draw[red] (0.153,55.30)  -- (0.146,54.10);
\end{axis}
\end{tikzpicture}
\caption{Comparison of model inference speed and CIDEr score on MSRVTT dataset. I, MV and Res refer to I-frame, motion vector and residual respectively. The test is run on 1 Card V100 machine with batch size set to 1.}
\label{fig:inference_speed_fig}
\end{figure}
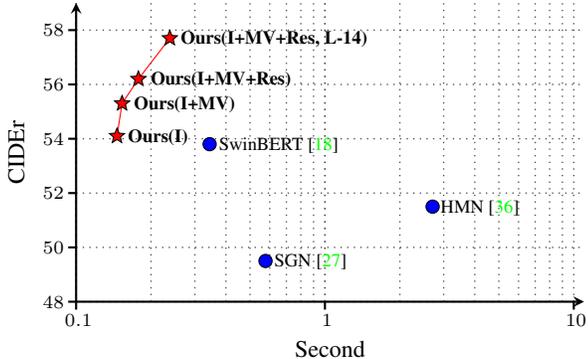

Video captioning requires both 2D appearance information, which reflects the objects in the video, and 3D action information, which reflects the actions. The interaction between these two types of information is crucial for accurately captioning the actions of objects in the video. Most of the existing methods~\cite{HMN,ORG-TRL,STG-KD} are shown in Fig.~\ref{fig:compare_with_prior_method} (the upper branch), mainly including the three-steps: (1) Decoding the video and densely sampling frames. (2) Extracting the 2D/3D features of the video frames offline. (3) Training the model based on these 2D/3D features. In these methods, densely sampled video frames result in significant redundancy, which in turn increases the computation and inference time of the model. This is because the model needs to extract features from each video frame and use all of these features as input. Furthermore, extracting 2D appearance features, 3D action features, and region features for each video frame requires additional time. To address the speed issue and improve inference speed, some recent works~\cite{SwinBERT,mvgpt} have adopted an end-to-end approach that avoids extracting multiple visual features offline. As shown in Fig.~\ref{fig:compare_with_prior_method} (The middle branch), the flow of their method is as follows: (1) Decoding the video and densely sample frames. (2) Take video frames directly as input and then end-to-end training model. These approaches involve a trainable visual feature extractor, rather than relying on multiple offline 2D/3D feature extractors. For example, SwinBERT~\cite{SwinBERT} uses VidSwin~\cite{vidswin} as the trainable feature extractor, while MV-GPT~\cite{mvgpt} uses ViViT~\cite{vivit}. While these two-steps methods address the time consumption associated with offline feature extraction, they do not alleviate the computational burden and time required to handle the redundancy of information.

To address the above problems, we propose an end-to-end video captioning method based on compressed video. Our work significantly simplifies the video caption pipeline by eliminating time-consuming video decoding and feature extraction steps.
As in Fig.~\ref{fig:compare_with_prior_method} (the lower branch), unlike previous methods, we take compressed video information as input and directly output a natural language description of the video. Compressed video is mainly composed of I-frame, motion vector and residual, and there is no redundant information between them, and they are all refined information. Therefore, the model needs less computation to process compressed domain information, and model inference is faster. At the same time, the end-to-end network structure in our proposed method can also avoid the time consumption caused by extracting multiple features. Besides, Our model is better at understanding the content of videos by utilizing the refined information in compressed domain, including the 2D feature from I-frame and the 3D action feature extracted from motion vector and residual. As shown in Fig.~\ref{fig:inference_speed_fig}, compared with other two-steps and three-steps methods, such as SwinBERT~\cite{SwinBERT}, HMN~\cite{HMN} and SGN~\cite{SGN}, our method is not only faster, but also has competitive performance. Our model comprises two parts, as depicted in Fig.~\ref{fig:main_framework}. One part consists of three encoders that extract features and an action encoder that fuses them, while the other part comprises a multimodal decoder that generates video captions. Specifically, we first extract the context feature, motion vector feature and residual feature of the compressed video through I-frame Encoder, Motion Encoder, and Residual Encoder, respectively. The context feature contains information about objects in the video, but action information is missing. In order to extract the action feature of the video, we fuse the motion vector feature, residual feature, and context feature through the action encoder. Then use the context feature and action feature as visual input of the multimodal decoder to generate video captions. 

The contributions of this paper are summarized below: 
\begin{enumerate}[leftmargin=*]{
\item We propose a simple and effective transformer that can take compressed video as input and directly generate a video description.
\item Our experimental results demonstrate that
our method is nearly 2× further than the fastest existing
state-of-the-art method in inference time, while maintaining competitive results on three challenging video captioning datasets, e.g., MSVD, MSRVTT and VATEX.
}
\end{enumerate}

\section{Related Work}

\noindent\textbf{Compressed vision task}. 
The main idea of introducing compressed video into current computer vision tasks is to utilizing the motion vector and residual on the compressed domain to avoid fully decode all frames from the video and save the storage space at the same time. 
Early work mainly base on MPEG-4 video codec~\cite{CoViAR,li2020slow,Huang_2021_CVPR,MM-ViT}. CoViAR~\cite{CoViAR} proposed a back-tracking technique to trace motion vectors back to I-frame, which works on MPEG-4. 
MM-ViT~\cite{MM-ViT} proposed a multi-modal transformer to process the I-frame, motion vector, residual and audio in the compressed video. 
Since the MPEG-4 codec is outdated, other works, e.g., MVCGC~\cite{huo2021compressed} and ATTP~\cite{huo2020lightweight} ,  is designed to work on other coedcs like H.264 and H.265 to ensure generalizability. 
Comparing with MPEG-4, H.264 and H.265 allow a more flexible yet complicated compression, which makes it more challenging to learn from compressed domain. 
MVCGC~\cite{huo2021compressed} proposed a self-supervised method to learn video representations by utilizing the mutual information between RGB video frames and motion vectors. ATTP~\cite{huo2020lightweight} designed a lightweight deep neural network to process the compressed video and achieve real time action recognition on embedded AI devices.
Similarly, our work is conducted on H.264 video codec, which is currently one of the most popular video codecs. 

\noindent\textbf{Video captioning.}
Video captioning aims to convert the content of videos into natural language descriptions, which requires the model to understand the objects in the video and the behavior of the objects. Some works focus on the design of the model structure. These methods usually extract features offline, and then models use these features to generate captions by designing different network architectures. HMN~\cite{HMN} proposed a hierarchical modular network that serves as a strong video encoder, which bridges videos and languages. ORG-TRL~\cite{ORG-TRL} proposes an object relational graph based encoder, which captures more detailed interaction features to enrich visual representation. SGN~\cite{SGN} designed a semantic grouping network to group video frames with discriminating word phrases of partially decoded caption. Some works explore additional information to help the model generate more accurate video captions. TextKG~\cite{textkg} propose a two-stream network capable of knowledge-assisted video description using knowledge graphs. Univl~\cite{univl} learns powerful vision-and-language representations by pre-training the models on large-scale datasets,~\eg, HowTo100M~\cite{how2100m} and WebVid-2M~\cite{webvid}. Some other works focus more on end-to-end video captioning generation. SwinBERT~\cite{SwinBERT} proposed an end-to-end transformer-based model, which takes video frame patches directly as inputs and then uses VidSwin to extract visual features. MV-GPT~\cite{mvgpt} designed an encoder-decoder model end-to-end to generate the video caption from video frames and transcribed speech directly.
We propose an end-to-end video captioning model based on the compressed domain without decoding video frames and extracting features offline, which not only accelerates the generation of captions, but also performs favorably against the state-of-the-art methods.

\section{Methods}

\begin{figure*}[tbp]
\centering
\includegraphics[width=0.95\textwidth]{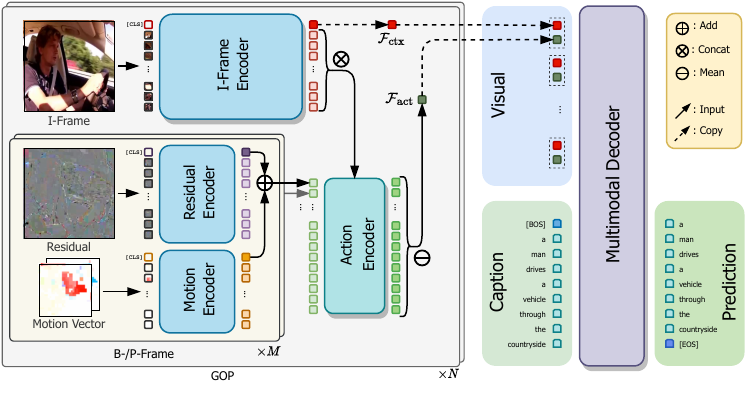}
\caption{The architecture of our proposed Compressed Video Captioner. \textit{Left}: The Compressed Video Transformer which extract video representation for each GOP. A large visual backbone is used to extract visual representations from I-frame, and two small Vision Transformer is used to extract residual and motion representations from compressed domain. After that, an action encoder is used to fuse the features. \textit{Right}: The Multimodal Decoder. We use a multimodal decoder with causal mask to learn caption.}
\label{fig:main_framework}
\end{figure*}

As mentioned above, our method aims to take the dense information (including I-frame, motion vector and residual) in compressed domain as input to accelerate inference and improve performance for video caption. 
To this end, we design an end-to-end transformer-based network as shown in Fig.~\ref{fig:main_framework}. In this section, we first detail the information in the compressed video in Sec.~\ref{section:3.1}, then introduce the model network in Sec.~\ref{section:3.2} and~\ref{section:3.3}, and finally introduce the training strategy of the model in Sec.~\ref{section:3.4}.

\subsection{The Structure of Compressed Video}
\label{section:3.1}

\begin{figure}[tbp]
    \centering
    \includegraphics[width=0.95\linewidth]{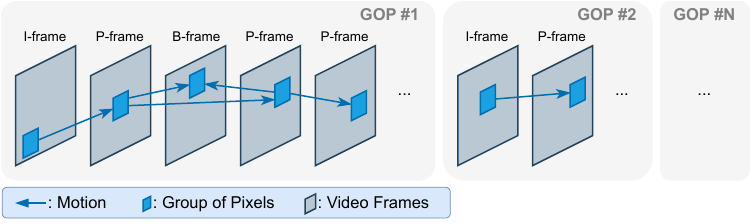}
    \caption{The GOP structure in compressed video. In each GOP, the first frame must be an I-frame, followed by several B/P-frames.}
    \label{fig:gop_structure}
\end{figure}

Modern video codecs utilizing the temporal redundancy of successive video frames to compress raw video. As shown in Fig.~\ref{fig:gop_structure}, most modern codecs (\eg, H.264, and H.265) divide video frames into three different types according to their dependencies with other frames: I-frame (intra coded frame), P-frame (predictive coded frame) and B-frame (bipredictive coded frame). 
I-frame is fully encoded independently using intra-prediction without relying on other frames. Other frames like B-frame and P-frame are encoded by referring to the other frames using inter-prediction, which is stored in the form of motion vector. 
Motion vector describes the movement of a group of pixels from source (reference frames) to destination (current B-frame or P-frame), which contains highly compressed motion information of successive video frames. 
The difference between P-frame and B-frame is that B-frame could refer to the frames before or after it, while P-frame only refer to the frames before it. 
Since predicting a frame using neighboring frames could be inaccurate, an additional residual error between the current frame and the prediction is calculated. 
We denote $\mathcal{I}_I$, $\mathcal{I}_P$ and $\mathcal{I}_B$ as decoded I-frame, P-frame, and B-frame, and $\mathcal{I}_{mv}$ and $\Delta_{res}$ as the motion vector and residual of P-/B-frame respectively.
In compressed domain, the P-frame and B-frame could be reconstructed by
\begin{equation}
    \mathcal{I}_{B/P} = \mathrm{Pred}(\mathcal{I}_{mv}, \mathcal{I}_{ref}) + \Delta_{res}
    \label{eq:reconstruct}
\end{equation}
where $\mathcal{I}_{ref}$ is the referenced frame, and $\mathrm{Pred}$ is the prediction method to reconstruct current frame based on motion vector and referenced frame. 
Since the reconstruction process is time consuming, our model takes highly compressed information from compressed domain directly as input to achieve end-to-end video captioning. 

Moreover, successive frames are divided into several groups, which is called Groups of Pictures (GOP). GOP is an independent encoding or decoding unit, which means that the frames in a GOP do not refer to any frames on other GOP. Each GOP starts with an I frame, followed by several P-frames or B-frames. 
For each GOP, we take one I-frame and $M$ B-/P-frames as inputs. The B-/P-frames are uniformly sampled from each GOP, and we only use their motion vector and residual as replacements. Therefore, the visual inputs of our model would be
\begin{equation*}
    \begin{split}
        X &= [\mathcal{I}_I^{(1)}, \mathcal{I}_{mv}^{(1,1)}, \Delta_{res}^{(1,1)}, \dots, \mathcal{I}_{mv}^{(M,1)}, \Delta_{res}^{(M,1)}], \\
        &\dots, [\mathcal{I}_I^{(N)}, \mathcal{I}_{mv}^{(1,N)}, \Delta_{res}^{(1,N)}, \dots, \mathcal{I}_{mv}^{(M,N)}, \Delta_{res}^{(M,N)}]
    \end{split}
\end{equation*}
where $N$ is the number of GOP sampled from each video and $M$ is the total number of P-/B-frames sampled from each GOP. We set $N$ according to the average GOP number, and $M$ is equal to the maximum number of P-/B-frames in each GOP. $M$ is equal to $\mathrm{KeyInt} - 1$, where $\mathrm{KeyInt}$ is a hyperparameter during the video encoding process.

\subsection{Model Architecture for Compressed Domain}
\label{section:3.2}

Based on the GOP structure mentioned above, we proposed a transformer based structure to utilizing the dense information from the compressed domain. Fig.~\ref{fig:main_framework} (left) shows the main framework of our proposed compressed video transformer. 
The model takes all information of the compressed video as inputs, including I-frame, motion vector and residual, while maintaining a fast inference speed.
Specifically, we use three different Vision Transformers~\cite{vit} (ViT) as encoder to extract the visual features for I-frame, motion vector and residual. 
We adopt a pretrained Vision Transformer as the encoder to extract the context feature from the I-frame:
\begin{equation*}
    \mathcal{F}_\mathrm{ctx}^{(n)} = \mathrm{Encoder}_\mathrm{I}(\mathcal{I}_I^{(n)}).
\end{equation*}
For each B-frame or P-frame, we get a motion vector and a residual from the compressed domain.
We use two lightweight Vision Transformers as encoders to extract features from motion vectors and residuals. The motion and residual features is added together to generate the B-/P-frame features $\mathcal{F}_\mathrm{BP}^{(m,n)}$:
\begin{equation*}
        \mathcal{F}_\mathrm{BP}^{(m,n)} = \mathrm{Encoder}_\mathrm{mv}(\mathcal{I}_{mv}^{(m,n)}) + \mathrm{Encoder}_\mathrm{res}(\Delta_{res}^{(m,n)}).
\end{equation*}
In this way, for each GOP we obtain $M$ B-/P-frame features 
\begin{equation*}
    \mathcal{F}_\mathrm{BP}^{(n)}=[\mathcal{F}_\mathrm{BP}^{(1,n)}, \dots, \mathcal{F}_\mathrm{BP}^{(M,n)}].
\end{equation*}

As motion vector and residual lack fine-grained context information, we use features from motion vector and residual as queries to retrieve the rich context information in RGB frames instead of simply fusing them. We employ action encoder to integrate the object information of I-frame into the action information of motion vector and residual, which takes B-/P-frame features in current GOP $\mathcal{F}_\mathrm{BP}^{(n)}$ and the context feature $\mathcal{F}_\mathrm{ctx}^{(n)}$ as input to generate the action feature $\mathcal{F}_\mathrm{act}^{(n)}$ of current GOP. The action encoder is constructed by $N_a$ sets of alternately stacked self-attention and cross-attention blocks. 

Specifically, the workflow of the action encoder is as follows. Firstly, according to the reconstruction process described in Eq.~\ref{eq:reconstruct}, we utilize the self-attention module fuse the temporal representation of successive frames to obtain $\mathcal{F}_\mathrm{att}^{(n)}$:
\begin{equation*}
\begin{split}
\begin{array}{ll}
    X=\mathcal{F}_\mathrm{BP}^{(n)} + \mathrm{Emb_p} + \mathrm{Emb_t},\vspace{1.5ex} \\
    Q=W_q*X, K=W_k*X, V=W_v*X,\vspace{1.5ex} \\
    \mathcal{F}_\mathrm{att}^{(n)} = \mathrm{SelfAttention}(Q, K, V), \\
\end{array}
\end{split}
\end{equation*}
where $\mathrm{Emb_p}$ is the positional embeddings, $\mathrm{Emb_t}$ is the type embeddings, and $W_q, W_k, W_v$ are learnable matrices. The type embeddings are added to distinguish B-frames and P-frames. 
And then we use the cross-attention to integrate the $\mathcal{F}_\mathrm{ctx}^{(n)}$ from I-frame into the $\mathcal{F}_\mathrm{att}^{(n)}$ from the motion vector and residual. Finally, the action feature $\mathcal{F}_\mathrm{act}^{(n)}$ 
\begin{equation*}
\begin{split}
\begin{array}{ll}
    Q'=W_q'*\mathcal{F}_\mathrm{att}^{(n)}, K'=W_k'*\mathcal{F}_\mathrm{ctx}^{(n)}, V'=W_v'*\mathcal{F}_\mathrm{ctx}^{(n)}, \vspace{1.5ex}  \\
    \mathcal{F}_\mathrm{att}^{(n)'} = \mathrm{CrossAttention}(Q', K', V'), \vspace{1.5ex} \\
    \mathcal{F}_\mathrm{act}^{(n)}=\mathrm{Mean}(\mathcal{F}_\mathrm{att}^{(n)'}), 
\end{array}
\end{split}
\end{equation*}
where $W_q', W_k', W_v'$ are learnable matrices and $\mathrm{Mean()}$ is a function that calculates the average feature.

\subsection{Multimodal Decoder for Video Captioning}
\label{section:3.3}

The context features $\mathcal{F}_\mathrm{ctx}^{(n)}$ and action features $\mathcal{F}_\mathrm{act}^{(n)}$ for each GOP are contacted to form the visual representation:
\begin{equation*}
\begin{array}{ll}
    \mathcal{V} = [\mathcal{F}_\mathrm{ctx}^{(1)}, \mathcal{F}_\mathrm{act}^{(1)}, \dots, \mathcal{F}_\mathrm{ctx}^{(N)}, \mathcal{F}_\mathrm{act}^{(N)}].
\end{array}
\end{equation*}

Then we design a multimodal decoder to predict the video captions based on the visual representation $\mathcal{V}$. The multimodal decoder is composed of $N_m$ masked self-attention modules stacked as shown in Fig.~\ref{fig:main_framework} (right) and the workflow is as follows:
\begin{equation*}
\begin{split}
\begin{array}{ll}
    \mathcal{T}_\mathrm{<t} = \mathrm{Embedding}(Y_\mathrm{<t}),\vspace{1.5ex}  \\
    \mathcal{X} = \mathrm{Concat}(\mathcal{V}, \mathcal{T}_\mathrm{<t}),\vspace{1.5ex} \\
    \mathcal{X}' = \mathcal{X} + \mathrm{Emb_p'} + \mathrm{Emb_t'},\vspace{1.5ex} \\
    Q'' = W_q''*\mathcal{X}', K'' = W_k''*\mathcal{X}', V'' = W_v''*\mathcal{X}',\vspace{1.5ex} \\
    h_\mathrm{t} = \mathrm{MaskedSelfAttention}(Q'', K'', V''),\vspace{1.5ex} \\
    p(y_\mathrm{t}|\mathcal{V}, \mathcal{T}_\mathrm{<t}) = \mathrm{softmax}(\mathrm{Linear}(h_\mathrm{t})),\vspace{1.5ex} \\
\end{array}
\end{split}
\end{equation*}
where $Y_\mathrm{<t}$ is the words generated in previous $t-1$ steps, $\mathrm{Embedding()}$ is a function that converts one-hot word vectors into word embeddings, $\mathrm{Emb_p'}$ is the positional embeddings, $\mathrm{Emb_t'}$ is used to distinguish different modality of inputs, $W_q'', W_k'', W_v''$ are learnable matrices and $y_\mathrm{t}$ is the prediction of current step. In the multimodal decoder, position embedding and type embedding is added to distinguish the order and type of features respectively.

\subsection{Optimization}
\label{section:3.4}

We train our model using the cross-entropy loss function. Given the ground-truth indices of previous (t-1) words and the visual representation $\mathcal{V}$, we can get the predictions of the current t-th word $y_{t}^*$. After that, the training loss is computed as 
\begin{equation*}
    \begin{array}{ll}
    L=-\sum_{t=1}^l \log p(y_t^*|y_{:t-1}^*, \mathcal{V}),
    \end{array}
\end{equation*}
where $y_{1:T}^*$ is the ground truth sequence and $l$ is the total length of predicted captions. Notably, we add the label smoothing to mitigate overconfidence in implementation.

\section{Experiments}

\begin{table*}[tbp]
    \centering
    \setlength{\tabcolsep}{4.5pt}
    \renewcommand{\arraystretch}{1.2}
    \begin{center}
    \resizebox{\linewidth}{!}{
    \begin{tabular}{c|cc|ccc|cccc|cccc}
    \hline
    \multicolumn{1}{c|}{\multirow{2}{*}{Method}} & \multirow{2}{*}{Decoding} & \multirow{2}{*}{E2E} & \multicolumn{3}{c|}{Features} & \multicolumn{4}{c|}{MSVD} & \multicolumn{4}{c}{MSRVTT} \\
    \multicolumn{1}{l|}{} & & & 2D Appearance & 3D Action & Object Detection & B4 & M & R & C & B4 & M & R & C \\ \hline
    SAAT~\cite{SAAT} & \checkmark & - & IncepResnetV2 & C3D & - & 46.5 & 33.5 & 69.4 & 81.0 & 39.9 & 27.7 & 61.2 & 51 \\
    STG-KD~\cite{STG-KD} & \checkmark & - & ResNet101 & I3D & FasterRCNN & 52.2 & 36.9 & 73.9 & 93.0 & 40.5 & 28.3 & 60.9 & 47.1 \\
    PMI-CAP~\cite{pmicap} & \checkmark & - & IncepResnetV2 & C3D & - & 54.6 & 36.4 & - & 95.1 & 42.1 & 28.7 & - & 49.4 \\
    ORG-TRL~\cite{ORG-TRL} & \checkmark & - & IncepResnetV2 & C3D & FasterRCNN & 54.3 & 36.4 & 73.9 & 95.2 & 43.6 & 28.8 & 62.1 & 50.9 \\
    OpenBook~\cite{openbook} & \checkmark & - & IncepResnetV2 & C3D & - & - & - & - & - & 42.8 & 29.3 & 61.7 & 52.9 \\
    SGN~\cite{SGN} & \checkmark & - & ResNet101 & C3D & - & 52.8 & 35.5 & 72.9 & 94.3 & 40.8 & 28.3 & 60.8 & 49.5 \\
    MGRMP~\cite{MGRMP} & \checkmark & - & IncepResnetV2 & C3D & - & 55.8 & 36.9 & 74.5 & 98.5 & 41.7 & 28.9 & 62.1 & 51.4 \\
    HMN~\cite{HMN} & \checkmark & - & IncepResnetV2 & C3D & FasterRCNN & 59.2 & 37.7 & 75.1 & 104 & 43.5 & 29 & 62.7 & 51.5 \\
    {\color[HTML]{C0C0C0}UniVL~\cite{univl}} & {\color[HTML]{C0C0C0}\checkmark} & {\color[HTML]{C0C0C0}-} & \multicolumn{3}{c|}{{\color[HTML]{C0C0C0}S3D}} & {\color[HTML]{C0C0C0}-} & {\color[HTML]{C0C0C0}-} & {\color[HTML]{C0C0C0}-} & {\color[HTML]{C0C0C0}-} & {\color[HTML]{C0C0C0}42.2} & {\color[HTML]{C0C0C0}28.8} & {\color[HTML]{C0C0C0}61.2} & {\color[HTML]{C0C0C0}49.9} \\ \hline
    SwinBERT~\cite{SwinBERT} & \checkmark & \checkmark & \multicolumn{3}{c|}{VidSwin} & 58.2 & 41.3 & 77.5 & 120.6 & 41.9 & 29.9 & 62.1 & 53.8 \\ 
    {\color[HTML]{C0C0C0}MV-GPT~\cite{mvgpt}}  & {\color[HTML]{C0C0C0}\checkmark} & {\color[HTML]{C0C0C0}\checkmark} & \multicolumn{3}{c|}{{\color[HTML]{C0C0C0}ViViT}} & {\color[HTML]{C0C0C0}-} & {\color[HTML]{C0C0C0}-} & {\color[HTML]{C0C0C0}-} & {\color[HTML]{C0C0C0}-} & {\color[HTML]{C0C0C0}48.9} & {\color[HTML]{C0C0C0}38.7} & {\color[HTML]{C0C0C0}64} & {\color[HTML]{C0C0C0}60} \\ \hline
    Ours & - & \checkmark & \multicolumn{3}{c|}{CLIP} & 55.9 & 39.9 & 76.8 & 113.0 & 43.1 & 29.8 & 62.7 & 56.2 \\
    Ours(ViT/L14) & - & \checkmark & \multicolumn{3}{c|}{CLIP} & \textbf{60.1} & \textbf{41.4} & \textbf{78.2} & \textbf{121.5} & \textbf{44.4} & \textbf{30.3} & \textbf{63.4} & \textbf{57.2} 
    \\ \hline
    \end{tabular}
    } 
    \end{center}
    \setlength{\belowcaptionskip}{-1cm}
    \caption{Comparison with state-of-the-art methods on the test split of MSVD and MSRVTT. Decoding means decoding video frames, and E2E means end-to-end training without offline feature extraction. For a fair comparison, we gray out models that pre-train on large-scale datasets.}
    \label{msvd-msrvtt-result}
\end{table*}

\begin{table}[tbp]
    \centering
    \setlength{\tabcolsep}{8pt}
    \renewcommand{\arraystretch}{1.0}
    \begin{center}
    \begin{tabular}{c|cccc}
    \hline
     & B4 & M & R & C \\ \hline
    NITS-VC~\cite{nits} & 20.0 & 18.0 & 42.0 & 24.0 \\
    VATEX~\cite{vatex} & 28.4 & 21.7 & 47 & 45.1 \\
    ORG-TRL~\cite{ORG-TRL} & 32.1 & 22.2 & 48.9 & 49.7 \\
    Support-set~\cite{patrick2020support} & 32.8 & 24.4 & 49.1 & 51.2 \\
    SwinBERT~\cite{SwinBERT} & \textbf{38.7} & \textbf{26.2} & \textbf{53.2} & \textbf{73} \\
    {\color[HTML]{C0C0C0}VideoCoCa~\cite{videocaco}} & {\color[HTML]{C0C0C0}39.7} & {\color[HTML]{C0C0C0}-} & {\color[HTML]{C0C0C0}54.5} & {\color[HTML]{C0C0C0}77.8} \\ \hline
    Ours & 31.4 & 23.2 & 49.4 & 52.7  \\
    Ours(ViT/L14) & 35.8 & 25.3 & 52.0 & 64.8 \\ \hline
    \end{tabular}
    \end{center}
    \setlength{\belowcaptionskip}{-1cm}
    \caption{Comparison with state-of-the-art methods on the test split of VATEX.  For a fair comparison, we gray out models that pre-train on large-scale datasets.}
    \label{vatex}
\end{table}

\begin{table*}[tbp]
    \setlength{\tabcolsep}{15pt}
    \renewcommand{\arraystretch}{1.0}
    \begin{center}
    \begin{tabular}{c|c|cc|c|c}
    \hline
    \multicolumn{1}{c|}{\multirow{2}{*}{Method}} & \multirow{2}{*}{Data Type} & \multicolumn{3}{c|}{Inference Time $\downarrow$} & \multirow{2}{*}{CIDEr $\uparrow$} \\
    \cline{3-5}
    \multicolumn{1}{l|}{} & & Feature Extraction & Model Time & Total &  \\ 
    \cline{3-4} 
    \hline
    SGN  & RGB Video Frames   & 303 ms   & 275 ms   & 578 ms   & 49.5 \\
    HMN  & RGB Video Frames   & 2,710 ms   & 108 ms   & 2,818 ms   & 51.5 \\
    SwinBERT & RGB Video Frames       & \multicolumn{2}{c|}{339 ms} & 339 ms & 53.8 \\
    \hline
    Ours & I-frame                    & \multicolumn{2}{c|}{146 ms} & \textbf{146 ms} & 54.1 \\
    Ours & I-frame+MV                 & \multicolumn{2}{c|}{153 ms} & 153 ms & 55.3 \\
    Ours & I-frame+MV+Res        & \multicolumn{2}{c|}{178 ms} & 178 ms & \textbf{56.2} \\
    \hline
    \end{tabular}
    \end{center}
    \caption{A detailed comparison of speed with other methods on the test split of the MSRVTT dataset. During the test, the model is running on a NVIDIA Tesla V100 GPU and the batch size is set to 1. The time cost is computed on the overall MSRVTT test split.}
    \label{table_speed}
\end{table*}

\subsection{Datasets}

    \noindent\textbf{MSRVTT}~\cite{MSRVTT}  is a generic video captioning dataset that comprises $10,000$ video clips, with each clip annotated with $20$ captions. On average, each video clip lasts about $15$ seconds. The standard split involves the use of $6,513$ clips for training, $497$ clips for validation, and $2,990$ clips for testing.
    
    \noindent\textbf{MSVD}~\cite{MSVD} contains $1,970$ videos, with each video clip having $40$ captions. The average duration of each video clip is around $10$ seconds. We adopt the standard split, which involves using $1,200$ videos for training, $100$ videos for validation, and $670$ videos for testing.
    
    \noindent\textbf{VATEX}~\cite{vatex} is a large-scale dataset which contains about $41,250$ video clips. The duration of each video clip is between $10$ seconds, and $10$ English captions are manually annotated per clip.  We use the official training set for training and evaluate the results using the public test set.

\subsection{Evaluation Metrics}

To evaluate the effectiveness of our approach, we use the standard metrics for video captioning: BLEU@4 (B4)~\cite{BLEU}, METEOR (M)~\cite{METEOR}, ROUGE (R)~\cite{ROUGE}, and CIDEr (C)~\cite{CIDEr}. Each metric provides a unique perspective on the quality of the generated captions. BLEU@4 evaluates sentence fluency, METEOR assesses semantic accuracy, ROUGE measures word order, and CIDEr evaluates the degree to which the caption conveys key information. By considering these different metrics, we can comprehensively evaluate the performance of our model.

\subsection{Implementation Details}

Our model is implemented using PyTorch, and to read motion vectors and residuals from the compressed video, we utilize the x264 library in FFmpeg. Before training and testing, the videos are resized to 240 on its smallest edge and compressed using the H.264 codec with $\mathrm{KeyInt}$ set to 60. For each video, we fixedly sampled 8 GOPs, each of which contains 1 I-frame, 59 motion vectors, and 59 residuals. The size of the I-frame and residual is $3*224*224$, and the size of the motion vector is $4*56*56$. We use Adam with initial learning rate of $\mathrm{1e-4}$, $\beta_1=0.9$, $\beta2=0.999$ and the warmup strategy is adopted in the training. 
The maximum length of the caption sentence is set to 22, which contains two special tokens, \eg, $[\mathrm{CLS}]$ token and $[\mathrm{EOS}]$ token. The feature dimension in each block is set to 768, and the number of heads in multi-head architecture is set to 12 for all layers.
The batch size is set to 64 and the training epochs to 20. The I-frame encoder has 12 layers and is initialized with pre-trained weights from the CLIP~\cite{clip} visual encoder, while the other encoders and the multimodal decoder are randomly initialized. The layers for the motion encoder, residual encoder and action encoder are 2, 2 and 1, respectively. Lastly, we set the hyperparameters $N_a$ and $N_m$ to 2 and 2.

\subsection{Performance Comparison with SOTA Methods}
In order to verify the effectiveness of the method, we evaluated the proposed model against state-of-the-art methods on three public benchmark datasets. 

\noindent\textbf{MSVD dataset.}
The evaluation results on the MSVD dataset are reported in Table~\ref{msvd-msrvtt-result} (left). We conducted experiments using two sizes of the I-frame encoder, namely $B/16$ and $L/14$, with the results reported in the article based on $B/16$, unless otherwise stated. Our method using the $L/14$ I-frame encoder achieves the best performance on all metrics, with only SwinBERT~\cite{SwinBERT} performing better than our method using $B/16$. Our approach stands out by being able to directly utilize compressed domain information and extract visual features in real-time. The result shows that our model can efficiently extract information from the refined compressed domain information.

\noindent\textbf{MSRVTT dataset.}
In the MSRVTT benchmark, our method outperforms other approaches in all metrics, as shown in Table~\ref{msvd-msrvtt-result} (right). Specifically, both the based on $B/16$ model and based on $L/14$ model achieve higher scores compared to other methods. In particular, our method achieves a CIDEr score of $56.2$ / $57.2$, which represents a significant improvement of $+2.4$ / $+3.4$. This result demonstrates that our approach can generate captions with higher semantic accuracy than other methods based on video decoding~\cite{CIDEr}. CIDEr is particularly effective at capturing human consensus, which makes our achievement in this metric even more impressive.

\noindent\textbf{VATEX dataset.}
Our method is evaluated on a large-scale dataset, as shown in Table~\ref{vatex}. We achieve the second-best results on all metrics, falling behind SwinBERT~\cite{SwinBERT}. Our approach involves extracting visual features using three Vision Transformer encoders, while the I-frame encoder is initialized with the pre-trained CLIP~\cite{clip} model on LAION-400M~\cite{400m}. In contrast, SwinBERT uses the VidSwin backbone~\cite{vidswin}, which is pre-trained on the Kinetic-600 dataset~\cite{kinetics}. It is worth noting that LAION-400M is a large image-text dataset, while Kinetics-600 is a video-text dataset, and VATEX dataset is a subset of Kinetics-600 videos. 
SwinBERT outperforms our method on VATEX due to its backbone pre-trained on Kinetics-600.

\subsection{Speed Comparison with the SOTA Methods}
To evaluate the speed of our method, we compared it to three representative methods, namely SGN~\cite{SGN}, HMN~\cite{HMN}, and SwinBERT~\cite{SwinBERT}, as reported in Table~\ref{table_speed}. SGN is a three-step method that first decodes video frames and densely sample, then extracts the 2D appearance and 3D action features based on ResNet101~\cite{resnet} and C3D~\cite{c3d} (consuming $303$~ms) offline, and finally uses the visual features as the input of the model (consuming $275$~ms). Therefore, the total time for SGN to generate a video caption is $578$~ms. HMN achieves the best results among the three-steps models, but it is relatively slow as it requires offline region feature extraction based on Faster RCNN~\cite{fasterrcnn} (consuming $2,520$~ms), leading to its total time of $2,818$~ms. SwinBERT, on the other hand, is an end-to-end method that does not extract multiple features offline, using only $339$~ms.

Compared to these methods, our proposed method does not require a dense sampling of video frames or the extraction of multiple features offline. As shown in Table~\ref{table_speed}, our baseline method only considers the I-frame of the entire video, achieving a CIDEr score of $54.1$ and a total time of $146$~ms. By integrating the motion vector, we improved the CIDEr to $55.3$, demonstrating that the action information in the motion vector helps the model generate captions. Furthermore, by incorporating residual information, the CIDEr score is further improved by $0.9$ to reach $56.2$. Although considering three inputs increases our total inference time, our speed is still nearly $2$ times faster than SwinBERT, $3$ times faster than SGN, and $15$ times faster than HMN.

\begin{table}[tbp]
    \centering
    \setlength{\tabcolsep}{5pt}
    \renewcommand{\arraystretch}{1.0}
    \begin{center}
    \begin{tabular}{ccc|c|cccc}
    \hline
    \multicolumn{3}{c|}{Input}  & \multicolumn{1}{c|}{Module}  & \multirow{2}{*}{B4} & \multirow{2}{*}{M} & \multirow{2}{*}{R} & \multirow{2}{*}{C} \\
    \multicolumn{1}{c}{$\mathcal{I}_{I}$} & \multicolumn{1}{c}{$\mathcal{I}_{mv}$} & \multicolumn{1}{c|}{$\Delta_{res}$} & \multicolumn{1}{c|}{En\_A} & & & & \\ \hline
    \checkmark  & - & - & - & 41.6 & 29.7 & 62.3 & 54.1  \\
    - & \checkmark & - & - & 27.3 & 21.6 & 52.6 & 19.4 \\
    - & - & \checkmark  & - & 23.9 & 20.5 & 51.0 & 13.0 \\
    \checkmark  & \checkmark & - & \checkmark  & 43.4 & 29.9 & 62.6 & 55.3 \\
    \checkmark  & - & \checkmark  & \checkmark  & 42.2 & 30.0 & 62.5 & 54.9 \\
    \checkmark  & \checkmark & \checkmark  & - & 42.1  &  \textbf{30.1} & 62.4 & 54.3 \\
    \checkmark  & \checkmark & \checkmark  & \checkmark  & \textbf{43.1} & 29.8 & \textbf{62.7} & \textbf{56.2} \\ \hline
    \end{tabular}
    \end{center}
    \caption{Ablation study of different input on the test subset of MSRVTT. The $\mathcal{I}_I$, $\mathcal{I}_{mv}$ and $\Delta_{res}$ mean decoded I-frame, motion vector and residual respectively. And the En\_A means the action encoder.}
    \label{ablation study}
\end{table}

\begin{figure*}[tbp]
    \centering
    \includegraphics[width=0.99\linewidth]{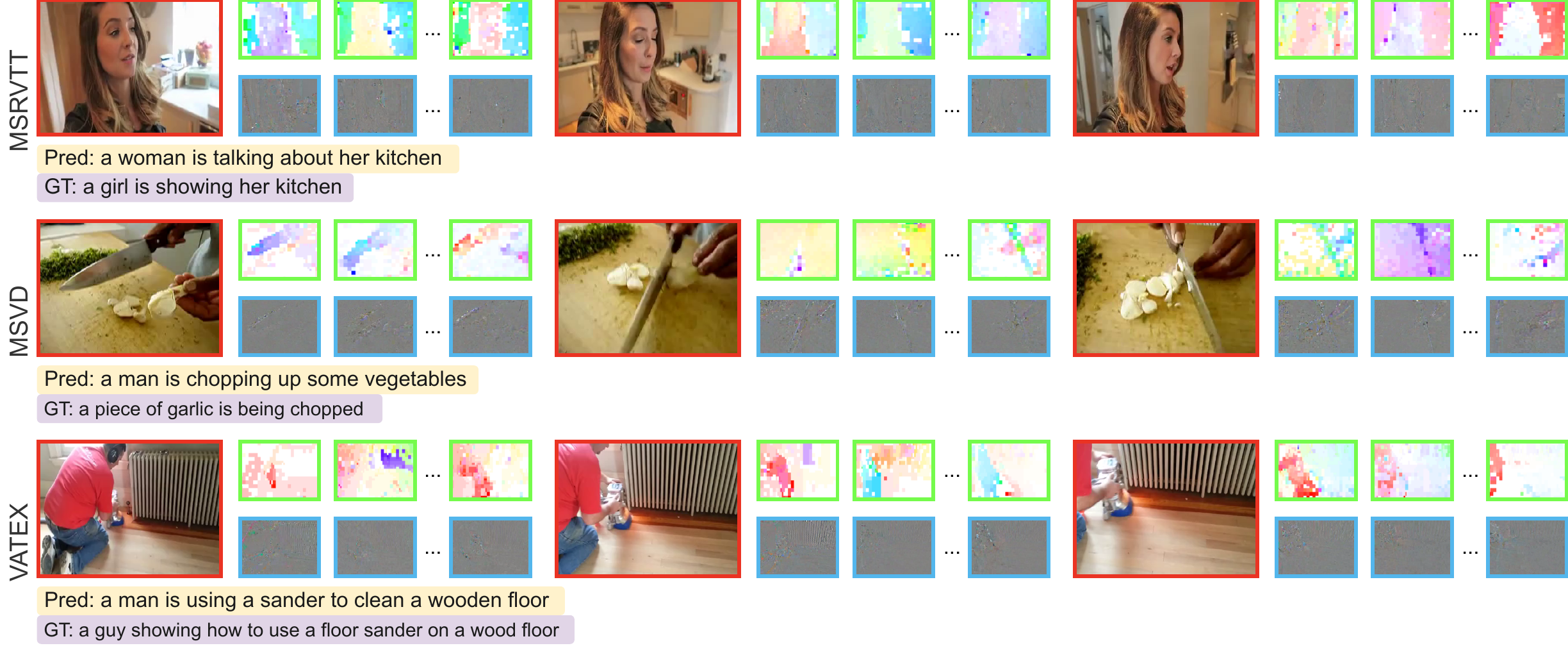}
    \caption{Qualitative results on the MSRVTT, MSVD and VATEX dataset. We show the input of our model, which is in compressed domain. The \textcolor{red}{red}, \textcolor{green}{green} and \textcolor{cyan}{blue} borders indicate I-frame, motion vector and residual, respectively.}
    \label{fig:qualitative_results}
\end{figure*}

\subsection{Ablation Study}

\noindent\textbf{Impact of input information.}
To evaluate the effectiveness of different input information in our method, we conducted several experiments on the MSRVTT dataset, as shown in Table~\ref{ablation study}. To investigate the role of I-frame, motion vector, and residual, we first experimented with using only one of them. As shown in Table~\ref{ablation study}, using only I-frame, motion vector, or residual achieved CIDEr scores of $54.1$, $19.4$, and $13.0$, respectively. This indicates that the model can directly use I-frame instead of motion vector and residual. By jointly using I-frame and motion vector and fusing their information through the action encoder, we achieved a CIDEr score of $55.3$. Similarly, using I-frame and residual achieved a score of $54.9$. This demonstrates that motion vector and residual can help the model generate more accurate captions. The performance of the model is further improved by inputting all three types of information, achieving a CIDEr score of $56.2$, an improvement of $1.7$. Removing the action encoder from the proposed method resulted in a slight drop in CIDEr scores, from $56.2$ to $54.3$. This demonstrates that the action encoder can help the model integrate the object information of I-frame into the action information of motion vector and residual.

\begin{table}[tbp]
    \begin{center}
    \setlength{\tabcolsep}{3pt}
    \renewcommand{\arraystretch}{1.0}
    \resizebox{\linewidth}{!}{
    \begin{tabular}{cc|c|cccc}
    \hline
    $\mathrm{KeyInt}$ & GOP Nums ($N$) & Inference Time & B4 & M & R & C \\ \hline
    250 & 2 & 153 ms & 39.6 & 28.7 & 60.8 & 49.5 \\
    60 & 2 & \textbf{131 ms} & 41.6 & 29.3 & 61.7 & 52.4 \\
    60 & 4 & 139 ms & 42.8 & \textbf{29.9} & 62.6 & 55.3 \\
    60 & 8 & 178 ms & \textbf{43.1} & 29.8 & \textbf{62.7} & \textbf{56.2} \\ 
    60 & 10 & 187 ms & 42.7 & 29.8 & 62.6 & 55.5 \\ \hline
    \end{tabular}
    }
    \end{center}
    \caption{Ablation study of GOP numbers on MSRVTT test subset.}
    \label{tab:abtation_gop}
\end{table}

\noindent\textbf{Impact of GOP numbers.} 
GOP is a fundamental unit in compressed video that affect the compression rate. A larger GOP size results in fewer GOP numbers and commonly higher compression rates. In video codec (\eg FFmpeg), the GOP size is determined by the KeyInt parameter. To investigate the impact of GOP size on our video caption model, we experimented with different GOP numbers and KeyInts, as shown in Table~\ref{tab:abtation_gop}. Comparing KeyInt values of $250$ and $60$, we observed that a smaller GOP size led to better model performance ($49.5$ CIDEr vs $52.4$ CIDEr). By sampling different GOP numbers under the same KeyInt, the best performance is achieved by setting GOP size to $8$ and KeyInt to $60$. While the performance
is improved with more GOPs, yet speed is decreased due to
increased computation as more information is included.

\begin{table}[tbp]
    \renewcommand{\arraystretch}{1.0}
    \begin{center}
    \begin{tabular}{ccccc|c}
    \hline
    En\_I & En\_M & En\_R & En\_C & De\_M & CIDEr \\ 
    \hline
    12 & 2 & 2 & 1 & 2 & 56.2 \\
    24 & 2 & 2 & 1 & 2 & 57.2 \\
    12 & 4 & 4 & 2 & 2 & 55.2 \\
    12 & 2 & 2 & 1 & 4 & 54.9 \\
    12 & 4 & 4 & 2 & 4 & 55.4 \\ 
    \hline
    \end{tabular}
    \end{center}
    \caption{Ablation study about module layers on the MSRVTT test subset. En\_I, En\_M, En\_R, En\_A and De\_M refer to the I-frame encoder, motion encoder, residual encoder, action encoder and multimodal decoder of the model respectively.}
    \label{ablation_layers}
\end{table}

\noindent\textbf{Impact of model layers.} To investigate the impact of different model layers on our proposed method, we conducted an ablation study on the MSRVTT test subset, as shown in Table~\ref{ablation_layers}. Giving that I-frame contains more complex information, we design a deep encoder with more layers for I-frame, while using a shallow encoder for motion vector and residual. Our results show that the performance of the model improves with an increase in the number of layers in the I-frame encoder ($56.2$ CIDEr to $57.2$ CIDEr). However, adding more layers to other modules did not result in further improvements in model performance.

\subsection{Qualitative Results}

As shown in Fig.~\ref{fig:qualitative_results}, we present the qualitative results of our proposed method on three datasets (\eg, MSVD, MSRVTT, and VATEX). Specifically, we visualize the input I-frame, motion vector, and residual and compare the predicted description to the ground truth. Our method consistently produces semantically consistent descriptions that closely align with the ground truth across all three datasets. Furthermore, the results demonstrate a superior ability to capture motion behavior in the videos.

\section{Conclusion}

In this paper, we introduce an end-to-end transformer-based model for video captioning that takes compressed video as input to eliminate redundant information. 
Our proposed method is evaluated on three challenging datasets and demonstrates that our proposed method is not only fast, but also competitive in performance with SOTA. In the future, we plan to further improve our method in two ways: (1) Add additional modalities such as audio, text, and knowledge graphs to enhance the quality of the generated captions. (2) Pre-train the model on a large-scale dataset to further boost the overall performance in compressed domain.

\section*{Acknowledgement} Libo Zhang was supported by Youth Innovation Promotion Association, CAS (2020111). Heng Fan and his employer received no financial support for research, authorship, and/or publication of this article. This work was done during internship at ByteDance Inc.

\clearpage

{
    \small
    \bibliographystyle{ieee_fullname}
    \bibliography{egbib}
}

\end{document}